\documentclass[conference]{IEEEtran}
\IEEEoverridecommandlockouts
\usepackage{cite}
\usepackage{amsmath,amssymb,amsfonts}
\usepackage{algorithmic}
\usepackage{graphicx}
\usepackage{textcomp}
\usepackage{xcolor}
\usepackage{multirow}
\usepackage{balance}
\usepackage{tikz}
\usepackage[utf8]{inputenc}
\def\BibTeX{{\rm B\kern-.05em{\sc i\kern-.025em b}\kern-.08em
    T\kern-.1667em\lower.7ex\hbox{E}\kern-.125emX}}

\newcommand\copyrighttext{%
	\footnotesize \textcopyright 2019 IEEE.  Personal use of this material is permitted.  Permission from IEEE must be obtained for all other uses, in any current or future media, including reprinting/republishing this material for advertising or promotional purposes, creating new collective works, for resale or redistribution to servers or lists, or reuse of any copyrighted component of this work in other works.}
\newcommand\copyrightnotice{%
	\begin{tikzpicture}[remember picture,overlay]
	\node[anchor=south,yshift=10pt] at (current page.south) {\fbox{\parbox{\dimexpr\textwidth-\fboxsep-\fboxrule\relax}{\copyrighttext}}};
	\end{tikzpicture}%
}

\begin{document}

\title{Double Transfer Learning for Breast Cancer Histopathologic Image Classification\\
}

\author{\IEEEauthorblockN{Jonathan de Matos\IEEEauthorrefmark{1}\IEEEauthorrefmark{4}, Alceu de S. Britto Jr.\IEEEauthorrefmark{2}\IEEEauthorrefmark{4}, Luiz E. S. Oliveira\IEEEauthorrefmark{3} and Alessandro L. Koerich\IEEEauthorrefmark{1}}\\ \IEEEauthorblockA{\IEEEauthorrefmark{1}École de Technologie Superiéure, Université du Québec, Montréal, QC, Canada\\
Email: jonathandematos@gmail.com, alessandro.koerich@etsmtl.ca}
\IEEEauthorblockA{\IEEEauthorrefmark{4}State University of Ponta Grossa (UEPG), Ponta Grossa, PR, Brazil}
\IEEEauthorblockA{\IEEEauthorrefmark{2}Pontifical Catholic University of Paraná, Curitiba, PR, Brazil\\
Email: alceu@ppgia.pucpr.br}
\IEEEauthorblockA{\IEEEauthorrefmark{3}Federal University of Parana, Curitiba, PR, Brazil\\ Email: luiz.oliveira@ufpr.com}
}

\maketitle
\copyrightnotice


\begin{abstract}
This work proposes a classification approach for breast cancer histopathologic images (HI) that uses transfer learning to extract features from HI using an Inception-v3 CNN pre-trained with ImageNet dataset. We also use transfer learning on training a support vector machine (SVM) classifier on a tissue labeled colorectal cancer dataset aiming to filter the patches from a breast cancer HI and remove the irrelevant ones. We show that removing irrelevant patches before training a second SVM classifier, improves the accuracy for classifying malign and benign tumors on breast cancer images. We are able to improve the classification accuracy in 3.7\% using the feature extraction transfer learning and an additional 0.7\% using the irrelevant patch elimination. The proposed approach outperforms the state-of-the-art in three out of the four magnification factors of the breast cancer dataset.
\end{abstract}

\section{Introduction}
Cancer is characterized by an uncontrolled growth and change in a group of cells. Such an anomaly is usually associated with a part of the body (e.g. breast, rectum). The mass formed by the uncontrolled growth is called tumor and can be classified as malign or benign. In a general definition, a benign tumor is not aggressive and does not need a rapid intervention because it does not destroy other cells and it does not spread quickly. On the other hand, a malign tumor can be referred to as cancer and presents a destructive, invasive and fast-spreading behavior. Therefore, it is necessary fast intervention to preserve the health of the patient \cite{cancer1}.

According to Torres et al. \cite{Torre2017}, breast cancer is the most prevalent type of cancer among women in 140 of 184 studied countries. Since 1990, the breast cancer mortality is decreasing due to the improvement in access to mammography exams and treatment. Although prevalent, in the United States the 5-year survival rate for breast cancer achieved 98\% for localized tumors due to early detection and adjuvant therapy \cite{cancer4}.

The breast cancer diagnosis starts with an imaging exam by X-ray (mammography), ultrasound or magnetic resonance \cite{cancer3}. Once found an anomaly on the images, the only way to confirm the presence and the type of tumor is by means of a biopsy. A biopsy consists of sampling the affected tissue followed by its analysis in a microscope. A pathologist can classify the sample in benign, malign or even the type of tumor. Biopsies have a high cost and are time-consuming due to the tissue extraction, preparation, and analysis by an experienced professional. In the United States 1.6 million women a year are submitted to biopsies, but only a quarter of the exams result in positive for malignant tumors. This factor increases the queue of exams, possibly delaying the start of the treatment for those that really require \cite{cancer7}. 
According to Gurcan et al. \cite{cancer6}, there is a growing demand for automated methods for diagnosing. Kalinli et al. \cite{cancer8} presented a comparative study of computer aided diagnosis (CAD) and pathologists' diagnosis in microscopic images of biopsies, called histopathologic images (HI). Gurcan et al. \cite{cancer6} raised questions related to inter-observer, intra-observer, physical, and psychological factors in the image analysis by pathologists. Inter-observer refers to the disagreement of malign and benign labeling of a sample by two pathologists. Intra-observer variations occur when a single pathologist observes the same sample within an interval of time, for example, of one day. His conclusions can be different between the two observations due to physiologic or psychological factors.

Considering the growing interest in CAD systems, Spanhol et al. \cite{breakhis} presented a study involving the recognition of the tumor type of breast cancer, while Kather et al. \cite{Kather2016} presented a study involving tissue recognition for colorectal cancer. Both works present an analysis of computational techniques to automate the recognition of tumors as well as provided access to their image datasets. Several works for breast cancer classification are based on monolithic \cite{Korkmaz20154026, Irshad2014390, Tashk20156165}. Peikari et al. \cite{Peikari20171078} used a cascade of SVMs for breast cancer classification. Valkonen et al. \cite{ISI:000404037600002} segmented whole slide images (WSI) of breast tissue using a Random Forest classifier. Balazsi et al. \cite{ISI:000383210600025} also studied the segmentation of WSI using a Random Forest and tessellation. {\color{black}{Araújo et al. \cite{Araujo2017} used a CNN with five convolutional layers and three fully-connected layers to classify breast cancer HI. They obtained 80.6\% of overall accuracy using the CNN as a classifier and 83.3\% using the same trained network as a feature extractor and a SVM with RBF kernel as a classifier.}}

Since 2012 when a Convolutional Neural Network (CNN) outperformed other traditional classification methods at the ImageNet contest \cite{alexnet}, deep approaches have also been proposed for breast cancer classification. However, the main constraints are the need of images of small size as input to limit the number of trainable parameters and the need for a large dataset to properly train such networks. Malon et al. \cite{Malon201297} proposed feature extraction from nuclei of HI to classify mitotic cells. Bayramoglu et al. \cite{7900002} proposed a custom CNN to classify breast cancer HI independent from magnification that uses low-level data augmentation to increase the size of the training dataset and image resizing. An approach using a CNN called BiCNN was proposed by Wei et al. \cite{wei} where they addressed the problem of a small size training set using data augmentation and transfer learning from the ImageNet dataset. Although producing good results, the CNN methods require data augmentation to produce more images for training as they are data-driven methods. On the other hand, the high number of samples increases the computational cost for training. Therefore, there is a trade-off between accuracy and cost.

Spanhol et al. \cite{breakhis2} proposed an interesting approach for data augmentation using a patching procedure that has the advantage of increasing the number of images while reducing their resolution to the input size required for a pre-trained CNN. Nevertheless, such an approach uses random patches and sliding windows. The main drawback is that HI of distinct classes may have large regions with similar information, for instance, an empty space or even the stroma. Therefore, a method for improving the quality of the patches could improve the accuracy of the model.

In this paper we propose the use of a source dataset to implement a transfer learning technique for filtering patches of a target dataset. Our main hypothesis is that filtering patches based on the tissue knowledge of a colorectal cancer dataset may provide a positive impact on classifying patches of a breast cancer dataset, due to the elimination of patches that do not contain relevant information, or even that do not contribute to distinguish between types of tumor. For such an aim, in the first step, we trained a SVM on the CRC dataset to classify input images as relevant or irrelevant tissue classes. This SVM will act as a filter in the next step. Besides that, we also transfer learning of deep convolutional representation from a pre-trained Inception-v3 CNN on ImageNet dataset. We also use a second feature extractor based on hand-crafted features called Parameter Free Threshold Adjacency Statistics (PFTAS) \cite{mahotas2013}. In the second step, we use the SVM trained on the CRC dataset to filter out irrelevant patches from our target dataset, before training a second SVM to the final task of classifying patches as malign or benign tumor. Furthermore, the patches can be aggregated to diagnose a patient. The first step allows us to increase the classification performance from 86.6\% to 90.3\% and the second step to further increase it to 91.0\%.

This paper is organized as follows. Section \ref{sec:prop} presents the proposed approach for filtering based on the transfer knowledge from the CRC dataset to the BreaKHis dataset, as well as the transfer representation-learning from pre-trained Inception-v3 CNN. Section \ref{sec:exp} presents the experimental results for filtering patches and classifying patients. In Section \ref{sec:disc} we discuss the results achieved by the proposed approach and compare them to the state-of-the-art. Finally, in the last section we present our conclusion and perspectives of future work.

\section{Proposed Method}
\label{sec:prop}
The main idea of the proposed approach is to transfer the knowledge from two source datasets to a target dataset where one of the source datasets has different data and different labels (different domain) and the other source dataset has similar data but different labels (same domain). The first dataset is ImageNet and it is used for feature representation transfer. The second dataset is CRC and it provides structural information (texture) about the tissue types of histopathologic images. Finally, the target domain is to classify histopathologic images from the BreaKHis dataset into two classes, malign and benign.

\begin{figure*}[htpb!]
	\centering
	\includegraphics[width=0.7\textwidth]{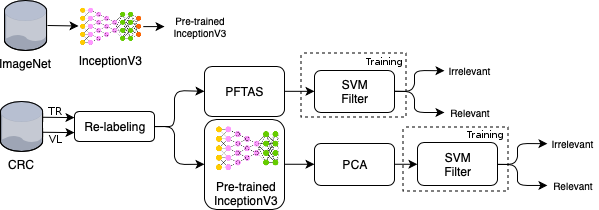}
	\caption{An overview of the double transfer learning from ImageNet and CRC datasets: feature extraction using two approaches (PFTAS and deep features from Inception-v3), filtering using CRC dataset as training set where TR is the training set and VL is the validation set.}
	\label{fig:filtering}
	\vspace{-0.1 in}
\end{figure*}

The proposed approach has two steps. In the first step, as illustrated in Fig.~\ref{fig:filtering}, we train a SVM to discriminate between relevant and irrelevant tissue structures in histopathologic images. In the second step, as illustrated in Fig.~\ref{fig:testing}, we use this SVM to filter images from the BreaKHis dataset before training another SVM whose aim is to detect benign or malign structures into the histopathologic images.

\subsection{Building a Filter}
\begin{figure}[htpb!]
\centering
{
\footnotesize
\setlength{\tabcolsep}{0.1em} 
\begin{tabular}{cccc}
  \includegraphics[width=0.24\columnwidth]{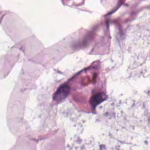} &   \includegraphics[width=0.24\columnwidth]{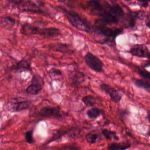} &   \includegraphics[width=0.24\columnwidth]{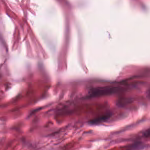} &   \includegraphics[width=0.24\columnwidth]{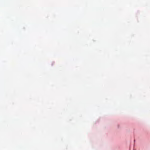} \\
  \includegraphics[width=0.24\columnwidth]{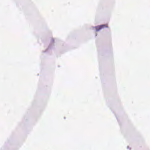} &   \includegraphics[width=0.24\columnwidth]{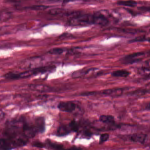} &   \includegraphics[width=0.24\columnwidth]{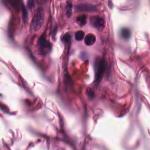} &   \includegraphics[width=0.24\columnwidth]{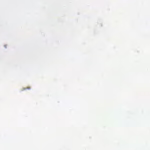} \\
  \vspace{0.08 in}
  (a) Adipose & (b) Complex Stroma & (c) Debris & (d) Empty \\
  \includegraphics[width=0.24\columnwidth]{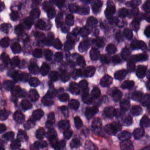} &   \includegraphics[width=0.24\columnwidth]{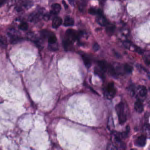} &   \includegraphics[width=0.24\columnwidth]{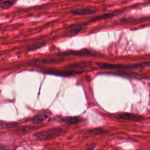} &   \includegraphics[width=0.24\columnwidth]{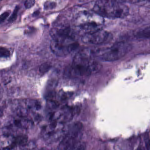} \\
  \includegraphics[width=0.24\columnwidth]{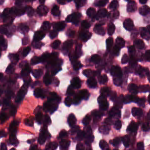} &   \includegraphics[width=0.24\columnwidth]{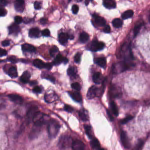} &   \includegraphics[width=0.24\columnwidth]{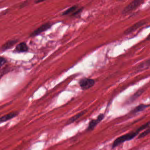} &   \includegraphics[width=0.24\columnwidth]{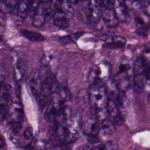} \\
 (e) Lympho & (f) Mucosa & (g) Stroma & (h) Tumor \\
\end{tabular}
}
\caption{Examples of 150$\times$150 patches from the CRC dataset}
\label{fig:crc_example}
\vspace{-0.2 in}
\end{figure}

The idea of building a filter is to train a SVM to classify histopathologic images (HI) as relevant or irrelevant as illustrated in Fig.~\ref{fig:filtering}. For such an aim, we use 
the CRC dataset which is composed of images of 150$\times$150 pixels labeled according to the structure they contain. Eight types of structures are labeled: Tumor (T), Stroma (S), Complex Stroma (CS), Immune or lymphoid cells (L), Debris (D), Mucosa (M), Adipose (AD), and Background or Empty (E). The total number of images is 625 per structure/tissue type, resulting in 5,000 images. Fig.~\ref{fig:crc_example} depicts examples of the images from the dataset.

The key to the filtering process is to define what is relevant and what is not in a histopathologic image. Intuitively, only background may be considered irrelevant, but we have evaluated other possible combinations of some tissues. We have based the tissue selection for relevant and irrelevant on the labeling of the classes of the CRC dataset. The tissues are labeled from the simplest to the most important for cancer diagnosis, taking into account that even Stroma and Complex Stroma tissues contain evidence of tumor. We could have evaluated all possible combination of tissues during the relabeling procedure, but this would produce $2^8$ experiments.

The first step consists in mapping the patches of the CRC dataset from eight to two categories. The aim of re-labelling the images is to handle the data imbalance among categories as well as to adapt it to the filtering task. Therefore, as shown in Table 	\ref{table:structurefilter}, we may have seven different scenarios where the total number of images mapped to our target categories, relevant (RE) and irrelevant (IR) categories, is always the same. The number of images of each source category is the same (e.g. 89 images of categories S, CS, L, D, M, AD, and E mapped to IR category for filter F1), but the ratio of the number of irrelevant images to the number of relevant images is imbalanced (e.g. 89 of S for irrelevant and 625 of T for relevant for filter F1). The imbalance is high for filters F1 and F7, but not so bad for filters F2 and F6, small for filters F3 and F5, while there is a balance for filter F4. This decision aims not to create a filter biased towards relevant or irrelevant classes.

\begin{table}[htpb!]
\renewcommand{\arraystretch}{1.3}
	\caption{Re-labeling of the images of the CRC dataset from eight categories (T, S, CS, L, D, M, AD and E) to two categories: relevant (RE) and irrelevant (IR). Numbers represent the number of images in each source and target category.}
	\label{table:structurefilter}
	\centering
	\begin{tabular}{|c|c|c|c|c|c|c|c|c|} 
		\hline
\bfseries Filter		&\bfseries T &\bfseries S &\bfseries CS &\bfseries L &\bfseries D &\bfseries M &\bfseries AD &\bfseries E \\
		\cline{2-9}
 \bfseries \#		&625 & 625 & 625 & 625 & 625 & 625 & 625 & 625 \\
		\hline
		\hline
		\multirow{3}{*}{\bfseries F1} &\multicolumn{1}{|c|}{RE} & \multicolumn{7}{c|}{IR} \\ \cline{2-9}
		&\multicolumn{1}{|c|}{625 } & \multicolumn{7}{c|}{625 } \\ \cline{2-9}
		& 625 & 89 & 89 & 89 & 89 & 89 & 89 & 89 \\
		\hline
		\hline
		\multirow{3}{*}{\bfseries F2} & \multicolumn{2}{|c|}{RE} & \multicolumn{6}{c|}{IR} \\ \cline{2-9}
		&\multicolumn{2}{|c|}{1,250} & \multicolumn{6}{c|}{1,248} \\ \cline{2-9}
		&625 & 625 & 208 & 208 & 208 & 208 & 208 & 208 \\
		\hline
		\hline
		\multirow{3}{*}{\bfseries F3} & \multicolumn{3}{|c|}{RE} & \multicolumn{5}{c|}{IR} \\ \cline{2-9}
		& \multicolumn{3}{|c|}{1,875} & \multicolumn{5}{c|}{1,875} \\
		\cline{2-9}
		& 625 & 625 & 625 & 375 & 375 & 375 & 375 & 375 \\
		\hline
		\hline
		\multirow{3}{*}{\bfseries F4} & \multicolumn{4}{|c|}{RE} & \multicolumn{4}{c|}{IR} \\ \cline{2-9}
		& \multicolumn{4}{|c|}{2,500} & \multicolumn{4}{c|}{2,500} \\
		\cline{2-9}
		& 625 & 625 & 625 & 625 & 625 & 625 & 625 & 625 \\
		\hline
		\hline
		\multirow{3}{*}{\bfseries F5} & \multicolumn{5}{|c|}{RE} & \multicolumn{3}{c|}{IR} \\ \cline{2-9}
		& \multicolumn{5}{|c|}{1,875} & \multicolumn{3}{c|}{1,875} \\
		\cline{2-9}
		& 375 & 375 & 375 & 625 & 625 & 625 & 625 & 625 \\
		\hline
		\hline
		\multirow{3}{*}{\bfseries F6} & \multicolumn{6}{|c|}{RE} & \multicolumn{2}{c|}{IR} \\ \cline{2-9}
		& \multicolumn{6}{|c|}{1,248} & \multicolumn{2}{c|}{1,250} \\
		\cline{2-9}
		& 208 & 208 & 208 & 208 & 208 & 208 & 625 & 625 \\
		\hline
		\hline
		\multirow{3}{*}{\bfseries F7} & \multicolumn{7}{|c|}{RE} & \multicolumn{1}{c|}{IR} \\ \cline{2-9}
		& \multicolumn{7}{|c|}{625} & \multicolumn{1}{c|}{625} \\
		\cline{2-9}
		& 89 & 89 & 89 & 89 & 89 & 89 & 89 & 625 \\
		\hline     
	\end{tabular}
	\vspace{-0.2 in}
\end{table}

For each scenario in Table \ref{table:structurefilter}, we extract two different feature sets to further train a two-class SVM: PFTAS and Inception-v3 deep features. The Parameter-Free Threshold Adjacency Statistics (PFTAS) is a handcrafted texture feature extractor which counts the black pixels in the neighborhood of a pixel (between zero and eight black pixels). A nine-bin histogram registers the total count for all the pixels in the image. The images are firstly binarized by the Otsu algorithm with three thresholds at each one of the channels (Red, Green, and Blue). Additionally, the thresholded images are bitwise inverted, doubling the number of values. This results in a 162-dimensional feature vector (9$\times$3$\times$3$\times$2). The second feature representation is generated by an Inception-v3 CNN pre-trained with ImageNet \cite{SzegedyVISW15}. The transfer representation learning uses the 2,048-dimensional fully-connected layer before the softmax layer as a feature vector. We used the Principal Component Analysis (PCA) method to reduce the feature vector dimensionality from 2,048 to a lower dimension. We have chosen four target dimensions to evaluate the impact of the reduction in the accuracy: 600, 400, 200 and 100 dimensions, where 600 dimensions correspond to 95\% of the accumulated variance from the most important components.

The aim of using two strategies for feature extraction is to verify if different features could improve the detection of irrelevant regions. Finally, a two-class SVM with RBF kernel is trained and evaluated for each resulting feature set for each scenario. Once the SVMs are trained, they are used for filtering the images of our target dataset, the BreakHis dataset.

\subsection{BreaKHis Patch Filtering}
At the first moment, our aim is not to classify all the structures present in the images of the BreaKHis dataset, but to improve the quality of the dataset before using it to train a classifier for our target task: given a HI, classify it as malign or benign. For such an aim, we use the SVMs trained at the first step, which will act as filters to eliminate irrelevant images from our target dataset.

\begin{figure}[htpb!]
\vspace{-0.1 in}
\centering
{
\footnotesize
\setlength{\tabcolsep}{0.1em} 
\begin{tabular}{cc}
  \includegraphics[width=0.44\columnwidth]{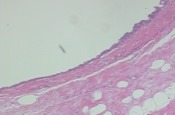} &   \includegraphics[width=0.44\columnwidth]{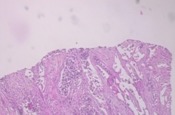} \\
  (a) Adenoma & (b) Ductal Carcinoma \\
\end{tabular}
}
\caption{Example of: (a) Adenoma (benign tumor) slide; (b) Ductal Carcinoma (malign tumor) slide}
\label{fig:adenoma_ductal_carcinoma}
\end{figure}

The proposed approach relies on the fact that some images of different classes contain a high amount of information in common. Fig.~\ref{fig:adenoma_ductal_carcinoma} depicts an example of two opposite types of tumor, an Adenoma (benign) and a Ductal Carcinoma (malign) that have a similar appearance in some areas which contains the slide background. Considering that these images will be patched into square regions and that there is no prior knowledge of each tissue types inside each image, but only what type of tumor an entire image contains, regardless if the tissue structure relevant to the diagnosis occupies a large or a small region within the image.

\begin{figure}[htpb!]
\vspace{-0.1 in}
\centering
{
\footnotesize
\setlength{\tabcolsep}{0.1em} 
\begin{tabular}{cccc}
  \includegraphics[width=0.24\columnwidth]{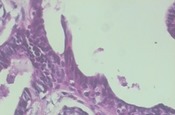} &   \includegraphics[width=0.24\columnwidth]{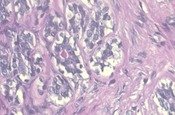} &
    \includegraphics[width=0.24\columnwidth]{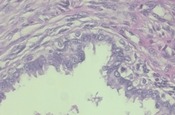} &   \includegraphics[width=0.24\columnwidth]{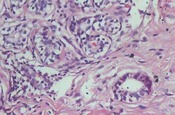}
    \vspace{-0.02 in}\\
  \includegraphics[width=0.24\columnwidth]{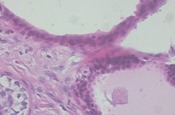} &   \includegraphics[width=0.24\columnwidth]{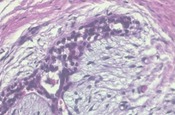} &
      \includegraphics[width=0.24\columnwidth]{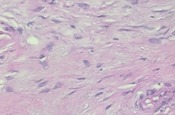} &   \includegraphics[width=0.24\columnwidth]{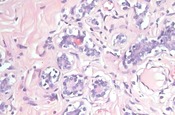} \\
      \vspace{0.07 in}
  (a) Adenoma & (b) Fibroadenoma & (c) Phyllodes & (d) Tubular\\

  \includegraphics[width=0.24\columnwidth]{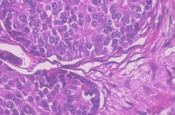} &   \includegraphics[width=0.24\columnwidth]{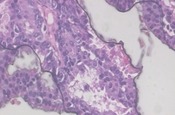} &
  \includegraphics[width=0.24\columnwidth]{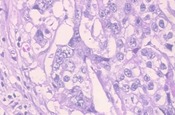} &   \includegraphics[width=0.24\columnwidth]{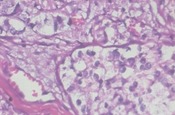}
      \vspace{-0.02 in}\\
  \includegraphics[width=0.24\columnwidth]{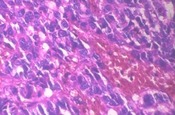} &   \includegraphics[width=0.24\columnwidth]{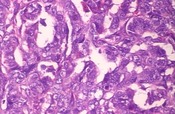} &
  \includegraphics[width=0.24\columnwidth]{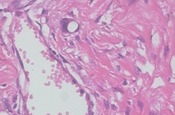} &   \includegraphics[width=0.24\columnwidth]{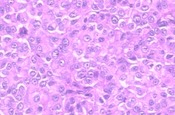} \\
    (e) Ductal & (f) Papillary & (g) Lobular & (h) Mucinous \\
\end{tabular}
}
\caption{Examples of images of the BreaKHis dataset: Benign tumors from (a) to (d). Malign tumors from (e) to (h).}
\label{fig:breakhis_example}
\end{figure}

BreaKHis is the target dataset that has its images or patients recognized as malign and benign breast tumors. This dataset introduced by Spanhol et al. \cite{breakhis} contains 7,909 HI images of tissues stained with Hematoxylin \& Eosin of 82 patients distributed in eight classes of tumors, where four classes are of malign tumors and the other four of benign tumors. The dataset is imbalanced by a factor of seven in the worst case, which means for example, that ductal carcinoma images (malign) have seven times more samples than the Adenosis (benign). Table \ref{table:classdistribution} shows the class distribution. Each image has 700$\times$460 pixels and for all patients there are images with four magnification factors: 40$\times$, 100$\times$, 200$\times$ and 400$\times$, which are equivalent to 0.49, 0.20, 0.10, 0.05 $\mu$m per pixel respectively. Images of different magnification were obtained from the minor to the major magnification factor of a microscope when the pathologist selects a region of interest of a biopsy specimen that is processed and fixed onto glass slides. Fig.~\ref{fig:breakhis_example} shows some examples of images from the BreaKHis dataset.

\begin{table}[htpb!]
\vspace{-0.1 in}
\renewcommand{\arraystretch}{1.2}
\caption{Image and patient distribution of BreaKHis dataset.}
\label{table:classdistribution}
\centering
\begin{tabular}{l l r r} 
		\hline
		\bfseries Tissue Type & \bfseries Tumor Type &  \bfseries \# of Images &  \bfseries \# of Patients \\
		\hline
		\multirow{5}{*}{{\rotatebox[origin=c]{90}{\parbox[c]{1cm}{\centering \bfseries Benign}}}} & Adenosis & 444 & 4 \\
		& Fibroadenoma & 1,014 & 10 \\
		& Phyllodes tumor & 453 & 3 \\
		& Tubular adenoma & 569 & 7 \\
		& Total & 2,368 & 24 \\
		\hline
		\multirow{5}{*}{{\rotatebox[origin=c]{90}{\parbox[c]{1cm}{\centering \bfseries Malign}}}} & Ductal carcinoma & 3,451 & 38 \\
		& Lobular carcinoma & 626 & 5 \\
		& Mucinous carcinoma & 792 & 9 \\
		& Papillary carcinoma & 560 & 6 \\
		& Total & 5,429 & 58 \\
		\hline
	\end{tabular}
	\vspace{-0.1 in}
\end{table}

\begin{figure*}[htpb!]
	\centering
	\includegraphics[width=0.8\textwidth]{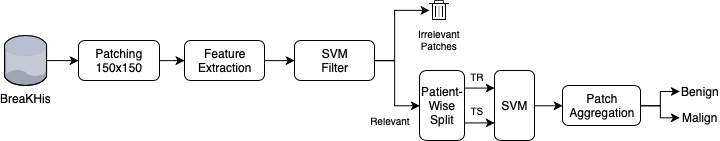}
	\caption{An overview of the proposed approach: patching, feature extraction (PFTAS or Inception-v3 + PCA), filtering by SVM, patient-wise splitting of relevant patches into training (TR) and test (TS) using the pre-defined folds, patch classification and aggregation using majority vote or sum rule.}
	\label{fig:testing}
\end{figure*}

First, the images of the BreaKHis dataset are patched by a 150$\times$150 rectangular sliding window with 3.3\% and 8\% of vertical and horizontal overlapping respectively. On average, a 700$\times$460 image produces 15 patches. Next, we use the same feature extraction approaches presented previously, namely the PFTAS and the Inception-v3 deep features, to extract features from the patches. These feature extractors work in parallel and independently of each other. For the sake of simplicity, in Fig.~\ref{fig:testing} we represent a single block of feature extraction. Besides, the PCA is also used to reduce the dimensionality of the feature vector produced by the Inception-v3. The corresponding SVM filter trained on CRC or Inception-v3 features is used to discard irrelevant patches from the BreaKHis dataset. Only the relevant patches are used then to train a two-class SVM to classify the patches of the BreaKHis dataset. 

The classification approach follows the protocol proposed by Spanhol et al. \cite{breakhis} that uses five folds with a patient-wise percentage of 70\% of patients for training and 30\% for testing. Furthermore, such a protocol ensures that all images of a patient are either in the training set or in the test set, but not in both sets at the same time. Therefore, the relevant patches are split using the same protocol and the two-class SVMs are trained in the training set with 5-fold cross validation to classify patches as Benign or Malign. At the end, we aggregate the patch classification results to come up to a whole image or patient classification.

\section{Experimental Results}
\label{sec:exp}
In the development phase, we trained the SVM filters with 85\% of the CRC dataset while 15\% of the images were used for validation. No test set was necessary because the purpose was not to evaluate the filter itself, but the filtering process. The selection between train and validation was random and stratified. In the development phase, we trained 35 SVM classifiers with RBF kernel using grid-search and 5-fold cross-validation for hyperparameter tuning on the data distribution shown in Table \ref{table:structurefilter}. Seven SVMs were trained with the PFTAS features and 28 SVMs were trained with the Inception-v3 deep features, seven for each PCA dimension (100, 200, 400 and 600). The SVM filters were optimized to achieve the best accuracy for classifying relevant/irrelevant images on the validation set. Table \ref{tab:filter_accuracy} presents the accuracy on the validation set for SVM filters trained on the CRC dataset.

\begin{table}[htpb!]
\vspace{-0.1 in}
\renewcommand{\arraystretch}{1.2}
\caption{Accuracy of the seven different filters for the validation set.}
\centering
\setlength{\tabcolsep}{0.45em} 
\begin{tabular}{lrrrrrrrr}
\hline
		& &\multicolumn{7}{c}{\bfseries Filter}\\
               \bfseries Feature & & \bfseries F1 & \bfseries F2 & \bfseries F3 & \bfseries F4 & \bfseries F5 & \bfseries F6 & \bfseries F7 \\
               \hline
PFTAS & & 95.2 & 87.5 & 92.7 & 95.1 & 96.8 & 98.9 & 99.5 \\
Inception-v3 + PCA 100 & & 97.9 & 94.7 & 96.8 & 98.0 & 98.0 & 99.5 & 100 \\
Inception-v3 + PCA 200&  & 97.9 & 94.1 & 97.0 & 98.0 & 98.0 & 99.5 & 100 \\
Inception-v3 + PCA 400 & & 97.9 & 94.7 & 97.0 & 97.7 & 97.9 & 99.5 & 100 \\
Inception-v3 + PCA 600 & & 98.4 & 94.7 & 96.8 & 98.1 & 97.9 & 99.5 & 100 \\
\hline
\end{tabular}
\label{tab:filter_accuracy}
\vspace{-0.1 in}
\end{table}

The 7,909 images of BreaKHis dataset generated 118,635 patches of dimension 150$\times$150 pixels. This size is the same as the images of the CRC dataset. However, keeping similar dimensions does not make the images from both datasets completely compatible, because pixels of BreaKHis' images represent 49 $\mu$m while pixels of the CRC dataset represent 74 $\mu$m in the best case. The pixel density problem cannot be solved because it is an acquisition parameter. The patches generated from the BreaKHis dataset have a vertical overlapping of five pixels and a horizontal overlapping of 12 pixels due to the distribution of five squared patches of dimension 150$\times$150 pixels over the 700$\times$460 image.

The accuracy per patient is calculated by Equation \ref{ref:eq1}, where $N_{\text{correct}}$ is the number of images correctly classified of one patient and $N_{\text{total}}$ is the total images of a patient.

\begin{equation}
\label{ref:eq1}
\textit{Patient Score} = \frac{N_{\text{correct}}}{N_{\text{total}}}
\end{equation}

The overall accuracy is obtained by Equation \ref{ref:eq2}, where \textit{Patient Score} comes from Equation \ref{ref:eq1} and \textit{Total Number of Patients} is the number of patients.

\begin{equation}
\label{ref:eq2}
\textit{Accuracy} = \frac{\sum \textit{Patient Score}}{\textit{Total Number of Patients}}
\end{equation}

All patches were submitted to the same filtering process, independently of their magnification factor. Table \ref{table:filtering} shows the number of patches, images, and patients selected as relevant after filtering and separated by the magnification factor.

\begin{table}[htpb!]
\vspace{-0.1 in}
\renewcommand{\arraystretch}{1.2}
	\caption{Percentage of patches, images and patients of BreaKHis dataset considered relevant after the filtering process with each filter using PFTAS features.}
	\label{table:filtering}
	\centering
	\begin{tabular}{c r r r r} 
		\hline
		\bfseries Magnification & \bfseries Filter & \bfseries Patches & \bfseries Images & \bfseries Patients \\
		\hline
		\multirow{7}{*}{40$\times$} & F1 & 48.7 & 10.6 & 1.7 \\
		& F2 & 100 & 73.0 & 31.5 \\
		& F3 & 95.1 & 62.7 & 24.1 \\
		& F4 & 98.7 & 75.4 & 33.5 \\
		& F5 & 100 & 99.5 & 81.3 \\
		& F6 & 93.9 & 88.4 & 70.1 \\
		& F7 & 100 & 96.9 & 90.9 \\
		\hline
		\multirow{7}{*}{100$\times$} & F1 & 39.0 & 6.1 & 0.8 \\
		& F2 & 96.3 & 60.3 & 21.3 \\
		& F3 & 93.9 & 40.7 & 10.4 \\
		& F4 & 97.5 & 51.7 & 14.7 \\
		& F5 & 100 & 95.8 & 68.7 \\
		& F6 & 91.4 & 88.5 & 70.2 \\
		& F7 & 100 & 97.3 & 93.7 \\
		\hline
		\multirow{7}{*}{200$\times$} & F1 & 47.5 & 9.4 & 1.3 \\
		& F2 & 100 & 61.5 & 21.3 \\
		& F3 & 92.6 & 32.8 & 7.6 \\
		& F4 & 97.5 & 41.9 & 10.1 \\
		& F5 & 100 & 87.4 & 53.3 \\
		& F6 & 97.5 & 89.0 & 59.7 \\
		& F7 & 100 & 99.8 & 97.3\\
		\hline
		\multirow{7}{*}{400$\times$} & F1 & 76.8 & 22.5 & 4.6 \\
		& F2 & 98.7 & 66.7 & 19.7 \\
		& F3 & 90.2 & 34.5 & 6.4 \\
		& F4 & 95.1 & 33.7 & 5.8 \\
		& F5 & 100 & 82.3 & 43.3 \\
		& F6 & 100 & 89.8 & 47.7 \\
		& F7 & 100 & 100 & 99.6 \\
		\hline
	\end{tabular}
\end{table}

After filtering the patches of the BreaKHis dataset, only the relevant ones are used in the next step. The relevant patches are split according to the protocol proposed by Spanhol et al. \cite{breakhis}. Therefore, five folds were used to train the SVM classifiers with RBF kernel. The best hyperparameters $C$ and $\gamma$ were found using cross validation and grid-search. We executed seven experiments, one for each filter. All filters were executed separately for each magnification factor. All 28 executions of filters with magnification factors were repeated five times, resulting in 140 executions. We executed one more batch of experiments without the filter, adding five more folds for each magnification factor adding more 20 experiments. These variations summed up to 160 experiments.

\subsection{Results for PFTAS Features}
Tables \ref{table:resultsnofiltering} and \ref{table:resultsfiltering} present the results of the first experiments at patch-level, image-level, and patient-level. The image-level results are obtained by aggregating patches using majority vote and the sum rule \cite{667881} on the probabilities predicted by the SVM classifier. The image classification results are used to calculate the patient-level accuracy. 

\begin{table}[htpb!]
\vspace{-0.1 in}
	\renewcommand{\arraystretch}{1.2}
	\caption{Mean percent accuracy and standard deviation for features extracted with PFTAS for the four magnification factors without filtering.}
	\label{table:resultsnofiltering}
	\centering
	\begin{tabular}{c c c c c c c} 
		\hline
		\bfseries  Magnif.& \bfseries Patches & \multicolumn{2}{c}{\bfseries Images} & \multicolumn{2}{c}{\bfseries Patients}\vspace{-3pt} \\
		& & \bfseries Sum & \bfseries Vote & \bfseries Sum & \bfseries Vote \\
		\hline
		40$\times$ & 82.9$\pm$3.4 & 85.0$\pm$4.4 & 85.4$\pm$4.4 & 86.1$\pm$4.5 & 86.4$\pm$4.7 \\
		100$\times$ & 83.0$\pm$3.8 & 84.9$\pm$3.9 & 84.6$\pm$3.9 & 86.6$\pm$4.9 & 86.3$\pm$4.9 \\
		200$\times$ & \textbf{86.5}$\pm$3.1 & \textbf{88.3}$\pm$3.5 & \textbf{88.5}$\pm$3.6 & \textbf{88.4}$\pm$4.6 & \textbf{88.7}$\pm$4.5 \\
		400$\times$ & 84.8$\pm$4.5 & 87.0$\pm$4.8 & 87.3$\pm$4.8 & 88.0$\pm$5.7 & 88.2$\pm$5.4 \\
		\hline
	\end{tabular}
\vspace{-0.1 in}	
\end{table}

\begin{table}[htpb!]
\vspace{-0.1 in}
	\renewcommand{\arraystretch}{1.2}
	\caption{Mean percent accuracy and standard deviation for features extracted with PFTAS for the four magnification factors using filter F7 (considering only the Empty or Background label of CRC dataset as irrelevant)}
	\label{table:resultsfiltering}
	\centering
		\begin{tabular}{c c c c c c c}
		\hline
		\bfseries  Magnif.& \bfseries Patches & \multicolumn{2}{c}{\bfseries Images} & \multicolumn{2}{c}{\bfseries Patients}\vspace{-3pt}  \\
		& & \bfseries Sum & \bfseries Vote & \bfseries Sum & \bfseries Vote \\
		\hline
			40$\times$ & 82.5$\pm$3.3 & 85.1$\pm$4.6 & 84.9$\pm$5.2 & 86.4$\pm$4.3 & 86.1$\pm$4.9 \\
			100$\times$ & 82.7$\pm$3.6 & 85.4$\pm$3.9 & 84.8$\pm$3.6 & 87.0$\pm$4.3 & 86.6$\pm$4.1 \\
			200$\times$ & \textbf{86.7}$\pm$3.6 & \textbf{89.0}$\pm$3.9 & \textbf{88.9}$\pm$4.3 & \textbf{89.2}$\pm$5.1 & \textbf{89.3}$\pm$5.3 \\
			400$\times$ & 84.8$\pm$4.6 & 87.0$\pm$4.9 & 87.2$\pm$5.0 & 87.9$\pm$5.7 & 88.2$\pm$5.8 \\
			\hline
		\end{tabular}
\end{table}
	
The results in Table \ref{table:resultsfiltering} consider only the patches filtered by filter F7, that considers as irrelevant only the Empty (or Background) images of CRC dataset because it produced the best results in terms of correct filtering (see Table \ref{tab:filter_accuracy}) and it never excluded patients (see Table \ref{table:filtering}). Only the magnification factor 200$\times$ provided improvements with filtering as highlighted in Table \ref{table:resultsfiltering}. However, the results of both Table \ref{table:resultsnofiltering} and \ref{table:filtering} surpasses the results presented in \cite{breakhis}. Spanhol et al. \cite{breakhis} used an SVM classifier and the PFTAS feature to achieve patient-wise mean percent accuracy of 81.6$\pm$3.0, 79.9$\pm$5.4, 85.1$\pm$3.1 and 82.3$\pm$3.8 for 40$\times$, 100$\times$, 200$\times$ and 400$\times$ magnification respectively.

\subsection{Results for Inception-v3 Deep Features}
\begin{figure}[htpb!]
\vspace{-0.1 in}
	\centering
	\includegraphics[width=3.5in]{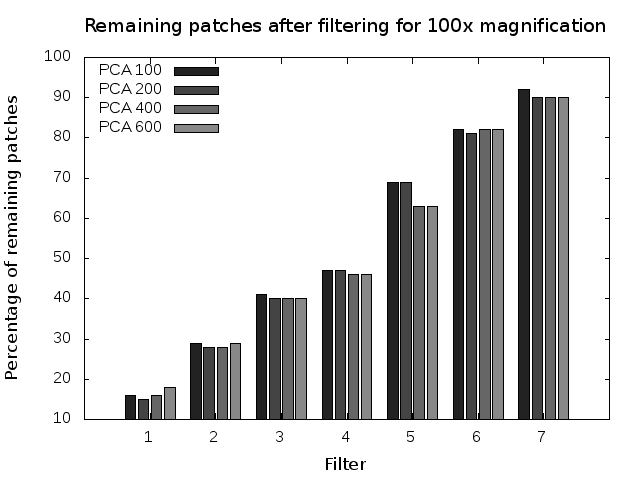}
	\caption{Percentage of remaining patches (relevant) for each filter using Inception-v3 deep features.}
	\label{fig:filter_statistics_tensorflow}
	\vspace{-0.1 in}
\end{figure}

Fig.~\ref{fig:filter_statistics_tensorflow} presents the results of the filtering process considering the percentage of remaining patches when the Inception-v3 deep features were used instead of PFTAS features. This graph represents only the remaining patches for the magnification factor 100$\times$. All other magnifications (40$\times$, 200$\times$, and 400$\times$) follow a very similar distribution. Aggressive filters such as filters F1 to F3, consider all images from a patient as irrelevant and in this case, we have discarded these filters. Besides that, some filters may exclude all patches from an image, but this is not a problem, meaning that the excluded image does not contribute to the classification.

Table \ref{tab:results_tensorflow} shows only the results at patient-level by aggregating patches using the sum rule and considering the two best filters (F6 or F7) or no filtering. In general, the mean percent accuracy for three out of four magnifications is better than those achieved with the PFTAS features. It is also possible to see that the filtering process only loses in the magnification factor of 40$\times$ for the feature vector of dimension 200.

\begin{table}[htpb!]
	\renewcommand{\arraystretch}{1.2}
	\caption{Mean percent accuracy and standard deviation for Inception-v3 deep features reduced by PCA and classified with an SVM classifier. The results are patient-wise aggregated by the sum rule. The best results for each magnification are in bold.}
	\label{tab:results_tensorflow}
	\centering
		\begin{tabular}{cccrr}
		    \hline
			\bfseries PCA & \bfseries Magnif. & \bfseries Filter & \multicolumn{1}{c}{\bfseries Filter}  & \multicolumn{1}{c}{\bfseries No Filter} \vspace{-3pt} \\
			\bfseries Dimension & & & \bfseries  &  \\
			\hline
			\multirow{4}{*}{ 100} & 40$\times$ & F6 & \bf 89.9$\pm$3.6 & 88.5$\pm$3.8 \\
			& 100$\times$ & F7 & \bf 91.0$\pm$3.0 & 90.3$\pm$3.4 \\
			& 200$\times$ & F7 & \bf 89.7$\pm$3.6 & 88.6$\pm$3.6 \\
			& 400$\times$ & F6 & 86.7$\pm$1.6 & 85.7$\pm$2.4 \\
			\hline
			\multirow{4}{*}{ 200} & 40$\times$ & F6 & 88.7$\pm$2.9 & 89.0$\pm$3.4 \\
			& 100$\times$ & F7 & 90.6$\pm$3.4 & 90.0$\pm$3.7 \\
			& 200$\times$ & F7 & 89.7$\pm$3.0 & 89.1$\pm$2.7 \\
			& 400$\times$ & F7 & 86.9$\pm$2.2 & 85.4$\pm$2.6 \\
			\hline
			\multirow{4}{*}{ 400} & 40 & F6 & 89.6$\pm$3.4 & 89.2$\pm$3.6 \\
			& 100$\times$ & F7 & 90.1$\pm$3.7 & 89.8$\pm$3.8 \\
			& 200$\times$ & F7 & 89.7$\pm$2.9 & 89.1$\pm$3.0 \\
			& 400$\times$ & F6 & \bf 87.1$\pm$1.5 & 85.3$\pm$1.7 \\
			\hline
			\multirow{4}{*}{ 600} & 40$\times$ & F6 & 89.5$\pm$3.4 & 89.3$\pm$3.5 \\
			& 100$\times$ & F7 & 89.8$\pm$3.4 & 89.7$\pm$3.7 \\
			& 200$\times$ & F7 & 89.7$\pm$3.2 & 89.1$\pm$2.9 \\
			& 400$\times$ & F7 & 86.9$\pm$1.8 & 85.6$\pm$2.0 \\
			\hline
		\end{tabular}
\vspace{-0.1 in}
\end{table}
	
\begin{figure}[!ht]
\vspace{-0.1 in}
	\centering
	\includegraphics[width=\columnwidth]{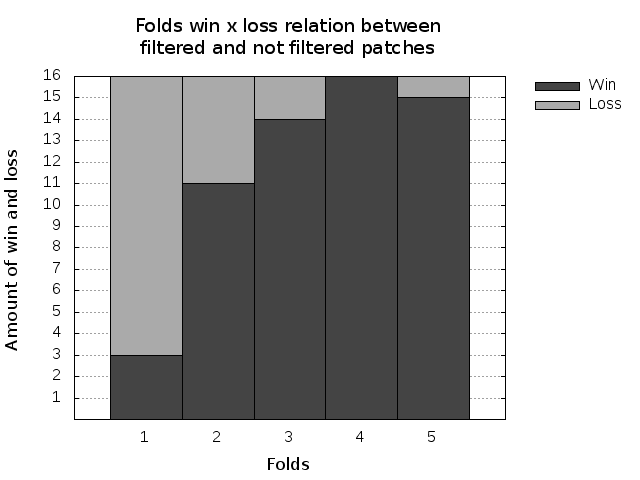}
	\vspace{-0.1 in}
	\caption{Comparing of losses and wins inside each folder for the best filtering and no filtering execution with Inception-V3 deep features.}
	\label{fig:loss_win}
	\vspace{-0.2 in}
\end{figure}

Fig.~\ref{fig:loss_win} shows the relation between wins and losses considering the filtered and non-filtered executions for each fold. We can notice that the filtering process produces worst results only in fold 1 for all 16 executions presented in Table \ref{tab:results_tensorflow}. For fold 1, it loses 13 times, but for the other folds, there are much more wins than losses. This result is related to the small number of patients for some categories of tumor, what makes some patients present in the test set of fold 1 and not present in the test set of other folds. There are only three patients with Phyllodes tumor. Therefore, when the fold is split into training (70\%) and test (30\%) only one patient remains in the test set. A similar situation occurs to the Adenosis and Lobular Carcinoma categories, which have only four and five patients respectively. Even if these two categories of tumor have a slightly higher number of patients, the imbalance also affects the performance. The images filtered in these test sets may not react well to the knowledge of the CRC dataset.

\section{Discussion}
\label{sec:disc}
The proposed filtering and patching approaches improved the classification results in comparison of using the whole image \cite{breakhis}. Table \ref{tab:results_comparison} presents a list of state-of-the-art methods that were evaluated on the BreaKHis dataset using similar experimental protocols. Spanhol et al. \cite{breakhis2} proposed a random patching procedure with the purpose of providing data augmentation for training a CNN, where random patches of sizes 64$\times$64 and 32$\times$32 pixels are generated by an overlapping sliding window. The number of patches generated by such a strategy is much higher than the proposed approach. However, our aim is not only augmenting the amount of data but also improving the patch quality by reducing the number of similar regions that may be present in different categories of tumor. Table \ref{tab:results_comparison} shows that, except for the 40$\times$ magnification, the proposed approach is the most accurate.

It is worth mentioning that the results presented by Han et al. \cite{csdcnn} outperform all the results reported in Table \ref{tab:results_comparison}, achieving 97.1\%, 95.7\%, 96.5\% and 95.7\% for the four magnification factors. However, such results are not comparable because they have used a completely different experimental protocol that did not respect the folder distribution proposed by Spanhol et al. \cite{breakhis}.

\begin{table}[ht!]
\vspace{-0.1 in}
\renewcommand{\arraystretch}{1.2}
\caption{Comparison with other state-of-art approaches on the BreaKHis dataset. Values represent percent accuracy.}
\label{tab:results_comparison}
	\centering
		\begin{tabular}{rcccc}
			\hline
			\bfseries Method & \multicolumn{4}{c}{\bfseries Magnification} \vspace{-3pt}  \\
			& \bfseries 40$\times$ &\bfseries 100$\times$ &\bfseries 200$\times$ &\bfseries 400$\times$ \\
			\hline
			Proposed (PFTAS) & 86.4 & 86.3 & 88.7 & \textbf{88.2} \\
			Proposed (PFTAS + Filter) & 86.1 & 86.6 & 89.3 & \textbf{88.2} \\
			Proposed (Inception-v3) & 88.5 & 90.3 & 88.6 & 85.7 \\
			Proposed (Inception-v3 + Filter) & 89.9 & \textbf{91.0} & \textbf{89.7} & 86.7 \\
			\hline
			Baseline \cite{breakhis} & 81.6 & 79.6 & 85.1 & 82.3 \\
			CNN (Alexnet) \cite{breakhis2} & 90.0 & 88.4 & 84.6 & 86.1 \\
			Deep Features (DeCaf) \cite{breakhis3} & 83.6 & 83.8 & 86.3 & 82.1 \\
			CNN + Fisher \cite{fisher} & 90.0 & 88.9 & 86.9 & 86.3 \\
			MI Approach \cite{breakhis4} & \textbf{92.1} & 89.1 & 87.2 & 82.7 \\
			\hline
		\end{tabular}
\end{table}
		
The main drawback of the proposed approach is the computational cost for training the SVM classifier due to the number of patches that we produced as well as due to dimensionality of the feature vectors. The number of instances impacts directly in the size of the kernel matrix and the dimensionality of the features impacts in the computation cost of the matrix calculation. The feature extraction is relatively time-consuming due to the size of the Inception-v3 CNN, but GPU computation helps to speedup this step. Although our approach is costly, the approaches that used only CNNs are much more expensive, taking more than 40 minutes for a single experiment, as reported in \cite{breakhis2}.

Another drawback of the proposed approach is the number of losses in fold 1, as shown in Fig.~\ref{fig:loss_win}. However, further analysis is required to detect what causes such losses in this particular fold. Therefore, we believe that there is still some room for an overall improvement if we could improve the performance for this fold.

\section{Conclusion}
\label{sec:concl}
We proposed a patch procedure with filtering to improve the quality of images for training a SVM classifier for breast cancer HI classification. The proposed approach uses two steps of transfer learning, one at feature extraction level, where deep features are extracted using an Inception-v3 CNN pre-trained with ImageNet dataset and the other at tissue structure level, where tissue images from another HI dataset are used to filter out irrelevant tissue structures of our target dataset. The experimental results have shown that the proposed filtering approach using transfer learning outperforms previous approaches. The proposed approach also helped in solving the problem of scarcity of data besides improving the quality of the patch images {\color{black}{due to the SVM classifier that learns from features extracted by other methods.}}

The results achieved by the proposed approach have shown that it is possible to exploit the knowledge from other datasets to improve the accuracy of a particular classification task. A common problem in HI is the lack of samples due to the expensive process of acquisition, analysis and labeling of images. Therefore, we used the knowledge acquired from the ImageNet dataset to extract meaningful features. Despite the differences between the images of the ImageNet and HI datasets, the Inception-v3 CNN performed well.

Another possible approach could be training from scratch using other HI dataset. However due to the low number of samples in such a kind of datasets, training a CNN is not feasible. To circumvent this problem, the fine-tuning is a way of reusing the representation learned from other tasks or datasets and adapt them to a new context by retraining only a small part of a CNN. We intend to use this approach in future works. The knowledge transferred from the CRC dataset was useful and it provided some improvement. In the future, we intend to use the activation layers of a CNN trained on other HI datasets to extract patches based on class-activated regions. Our idea is that activation regions with higher values have a strong contribution to the final classification.

\balance

\bibliographystyle{IEEEtran}
\bibliography{IEEEabrv,sample}

\end{document}